\pdfoutput=1
\documentclass[conference]{IEEEtran}
\usepackage{import}
\usepackage[subpreambles,mode=image]{standalone}

\usepackage{times}
\usepackage[utf8]{inputenc}

\usepackage[numbers]{natbib}

\usepackage{microtype}
\usepackage{graphicx}
\usepackage{epstopdf}
\graphicspath{{figures/}}
\DeclareGraphicsExtensions{.eps,.pdf,.jpeg,.png}

\usepackage{array}
\usepackage{float}
\usepackage{booktabs}
\usepackage[caption=false]{subfig}
\usepackage{floatflt} % Put figures aside a text
\newcolumntype{L}{>{$}l<{$}} % math-mode version of "l" column type
\newcolumntype{R}{>{$}r<{$}} % math-mode version of "r" column type
\let\labelindent\relax
\usepackage[inline]{enumitem}
\usepackage{xcolor}

\usepackage[hyphens]{url}
\usepackage[bookmarks=true]{hyperref}

\usepackage{notation}

\usepackage{amsthm}
\theoremstyle{definition}
\newtheorem{definition}{Definition}
\newtheorem*{definition*}{Definition}
\newtheorem{theorem}{Theorem}
\newtheorem{corollary}{Corollary}

\newtheorem{problem}{Problem}

\theoremstyle{plain}

\theoremstyle{remark}

\newtheorem*{remark}{Remark}
\newtheorem*{note}{Note}

\def\problemname{Problem}

\makeatletter
\renewcommand\paragraph{\@startsection{paragraph}{4}{\z@}%
                     {-12\p@ \@plus -4\p@ \@minus -4\p@}%
                     {-0.5em \@plus -0.22em \@minus -0.1em}%
                     {\normalfont\normalsize\bfseries}}
\makeatother

\usepackage[obeyFinal]{todonotes}
\setuptodonotes{inline}

\begin{document}
\title{Model-Free Reinforcement Learning for Symbolic Automata-encoded Objectives}

\author{%
\authorblockN{%
Anand Balakrishnan\authorrefmark{1},
Stefan Jak\v{s}i\'{c}\authorrefmark{2},
Edgar A.~Aguilar\authorrefmark{2},
Dejan Ni\v{c}kovi\'{c}\authorrefmark{2}, and
Jyotirmoy V.~Deshmukh\authorrefmark{1}
}
\authorblockA{\authorrefmark{1}%
  University of Southern California, Los Angeles, California, USA\\
  Email: \{anandbal, jdeshmuk\}@usc.edu}
\authorblockA{\authorrefmark{2}%
  AIT Austrian Institute of Technology GmbH, Vienna, Austria\\
  Email: \{stefan.jaksic, edgar.aguilar, dejan.nickovic\}@ait.ac.at}%
}

\maketitle
% \IEEEpeerreviewmaketitle%

\begin{abstract}
  Reinforcement learning (RL) is a popular approach for robotic path planning 
  in uncertain environments. However, the control policies trained for an
  RL agent crucially depend on user-defined, state-based reward functions. 
  Poorly designed rewards can lead to policies that do get maximal rewards
  but fail to satisfy desired task objectives or are unsafe. There
  are several examples of the use of formal language such as 
  temporal logics and automata to specify high-level task specifications
  for robots  (in lieu of Markovian rewards). Recent efforts have focused 
  on inferring state-based rewards
  from formal specifications; here, the goal is to provide (probabilistic) 
  guarantees that the policy learned using RL (with the inferred rewards) 
  satisfies the high-level formal specification. A key drawback of several
  of these techniques is that the rewards that they infer are sparse:
  the agent receives positive rewards only upon completion of the task
  and no rewards otherwise. This naturally leads to poor convergence 
  properties and high variance during RL. 
 
  In this work we propose using formal specifications in the form of symbolic 
  automata: these serve as a generalization of both bounded-time temporal
  logic based specifications as well as automata. Furthermore our 
  use of symbolic automata allows us to define non-sparse potential-based
  rewards which empirically shape the reward surface, leading to better
  convergence during RL. We also show that our potential-based rewarding
  strategy still allows us to obtain the policy that maximizes the satisfaction
  of the given specification.
  
\end{abstract}

% \input{sections/intro}
% \input{sections/prelims}
% \input{sections/problem}
% \input{sections/product}
% \input{sections/experiments}
% \input{sections/related}
% \input{sections/conclusion}

% !TEX root = ../main.tex
\section{Introduction}
\label{sec:intro}

Reinforcement learning (RL) is a paradigm for automatically synthesizing controllers
that perform a certain task through repeated interaction with the
environment~\cite{barto1983neuronlike,sutton2018reinforcement}.
In the standard RL setting, instead of explicitly programming an agent to perform the
task -- which is often infeasible due to task complexity -- a reward function is
specified for the actions an agent might take.
Thus, by taking actions in an environment and collecting rewards, an agent learns which
actions maximize the overall expected reward.

With the advent of deep neural networks in the last decade, associated to an
unprecedented increase in computational power, RL was able to solve incredibly complex
tasks using simple reward functions~\cite{mnih2015humanlevel,mnih2016asynchronous,silver2014deterministic,schulman2017proximal}.
This major accomplishment of \emph{deep reinforcement learning} was demonstrated in
various environments, from video games to robot locomotion.
A major ingredient in the success of RL is a \emph{well-designed reward function} that
encodes the task correctly.
Engineering a good reward function is still considered to be an art.

A good reward function must ensure that the agent will be able to learn how to reach
desired objectives according to designer intentions.
A poorly designed reward function can result in a situation where an agent learns to
maximize total rewards without actually satisfying the high-level objective intended by
the designer.
This situation is known as \emph{reward hacking}~\cite{amodei2016concrete}.
Reward hacking can lead to undesirable behaviors, which can be of particular importance
in the context of safety-critical systems, where failures are not acceptable.
To avoid reward hacking, there is a need for more principled approaches to reward
engineering.
%\emph{Reward hacking} refers to the situation where 
%A poorly designed reward function can lead to behavior that is undesirable in the best
%case, and fatal in the worst case.
%This class of issues are of particular importance in the context of safety-critical or
%expensive systems, where any failure mode is unacceptable.

Traditional reward functions are Markovian, i.e. they do not depend on history
to complete a given task.
This makes them inappropriate for history-aware objectives, such as sequential tasks,
where individual objectives must be completed in order, and security surveillance,
where an area needs to be patrolled within a time bound.

Techniques from formal methods can be effectively used to address the problem
of principled history-aware reward engineering.
For instance, automata-based approaches~\cite{sadigh2014learning,hasanbeig2018logicallyconstrained,hahn2020reward,lavaei2020formal}
have been used to solve the issue of non-Markovian objectives for
\emph{omega-regular} objectives for infinite behavior.
However, these methods are susceptible to the definition of \emph{sparse reward functions}.
Sparse reward functions suffer from long convergence because the agent has to
perform a long sequence of actions before seeing any concrete reward.
Finally, techniques that use quantitative semantics of temporal logics have shown to be highly
effective in finding optimal policies up to some finite control
horizon~\cite{aksaray2016qlearning,balakrishnan2019structured}.
Unfortunately, such techniques require that some history is stored to remove the non-Markovian
nature of the objective.
These approaches are also limited by the ``horizon'' of the tasks at hand, and scale
poorly to tasks that require a large history.
% \dn{I am missing here a sentence on what the TL approach is not able to do,
% but that we do instead with our SA.}

\subsection{Main Contributions}

In this paper, we propose a novel approach to encode a finite sequence of tasks using
symbolic automata~\cite{dantoni2017power}.
Symbolic automata enable the encoding of history-dependent goals, while providing rich
quantitative information about the means to achieve the overall objective.
We propose a reward function that uses both the automaton structure and its symbolic
information to accelerate the discovery of the optimal policy.
Our reward function consists of two components: (1) a sparse base reward that is given
only when the agent reaches the overall objective, and (2) an additive potential
function that shapes the base reward to ensure progress towards intermediate goals and
the final objective.
To the best of our knowledge, we present the first method of using symbolic automata as
task specifications for reinforcement learning.

We prove the soundness of our reward function on a \textit{product transition system} in
a model-based RL setting, to show that it solves the problem of maximizing the
probability of satisfying SA specification in an MDP.
Then, we empirically evaluate it in a model-free RL environment, where we use the
Q-learning algorithm and our reward to tackle three different problems: (1) goal
reachability, (2) recurrence, and (3) sequential tasks.
We demonstrate that our reward function indeed optimizes the learning process towards
the goals specified by the symbolic automaton specification.
This is done by showing a significant increase in the convergence rate when compared to
baseline rewards.

\subsection{Organization}

The rest of the paper is organzed as follows: Section~\ref{sec:prelims} provides
the theoretical prerequisites of our work.
In Section~\ref{sec:problem_s} we formalize the problem of policy synthesis for a
symbolic automata encoded task in a given MDP, for which we propose a RL rewarding
strategy in  Section~\ref{sec:rewarding_strategy}.
We empirically evaluate our reward strategy in three distinct scenarios in
Section~\ref{sec:experiments}.
We finalize our paper with an overview of the related work in
Section~\ref{sec:related} and conclude the paper in
Section~\ref{sec:conclusion}.

% !TEX root = ../main.tex
\section{Preliminaries}%
\label{sec:prelims}

%Given an \(n\)-dimensional problem state space \(D^n\), where \(D
%\subset \Re\) is a finite, valuation set for each dimension in \(D^n\).
%Let \(X =
%\Set*{x_1, \ldots, x_n}\) be a set of variables representing each dimension in \(D^n\).
\newcommand{\ignore}[1]{}

\subsection{Symbolic Automata}

Let \(X = \Set*{x_1, \ldots, x_n}\) be a set of variables, where each
$x_i$ takes values in some compact set $D \subseteq \Re$.
Let \(v: X \to D\) be a \emph{valuation} function (or just valuation) that maps a
variable \(x \in X\) to the value of  \(x\).
For example if $X = \Set*{x,y}$, then the function $v: x\mapsto 3, y\mapsto 5$ defines
the values of $x$ and $y$ to
$3$ and $5$ respectively.
Given the set of variables $X$, we can abuse notation and use
$v(X)$ to denote $(v(x_1),\ldots,v(x_n))$.
In other words $v(X)$ is some value in $D^n$.
We can define a metric over this space, for example, consider the Manhattan distance
between two valuations defined as follows:
\begin{equation}
  \label{eq:man}
  d_{man}(v,v') = \sum_{i=1}^n |v(x_i) - v'(x_i)|.
\end{equation}
If $v(X) = (3,5)$ and $v'(X) = (2,1)$, then $d_{man}(v,v') = 1 + 4 =5$.

\ignore{
  For example, let \(X = \{x, y\}\) be a set of variables, and let \(s = (2, 5)\)
  represent a vector in some 2-dimensional space \(S\).
  Then, a possible valuation function is:
  \begin{align*}
    v(s = (2,5), x) & = 2 \\
    v(s = (2,5), y) & = 5
  \end{align*}
}
% Let \(d: S \times S \to \Re_{\geq 0}\) be a distance associated with the metric space
% \(S\).

% Let $X = \{ x_1, \cdots, x_n \}$ be a set of real-valued
% variables defined over some bounded domain $D \subseteq \Re$,
% and $v~:~X \to D$ be a \emph{valuation} function.
% \jyo{This should be $X \to D^n$.
%   Also do we need to define $X$ as a metric space.
% }
% \dn{if we define our valuation as a function than it is $D$ and not $D^n$. it maps
%   each variable to a value, and not to a set of values}
% \jyo{My bad, maybe what I want to say is it should be $D \subseteq \Re^n$?}

% Let \(D \subseteq \Re\) be a metric space with some distance \(d\), and \(X\) be a set
% of variables defined over
% \(D\).
\begin{definition}[Predicate]
  A \emph{predicate \(\psi\) over \(X\)} is defined with the following recursive
  grammar: \[\psi := \top \mid \bot \mid x \sim k \mid \neg \psi \mid \psi \land \psi \]
  where \(x \in X, k \in D, \text{ and } \sim \in \Set{<, \leq, >, \geq, =}\).
  We denote by \(\Psi(X)\) the set of all predicates over \(X\).
\end{definition}

Given a valuation function \(v: X \to D\), we define the semantics of \(\psi\)
in terms of a satisfaction relation \(v \models \psi\) as follows:
\begin{center}
  \begin{tabular}{L  @{\ $\iff$\ } L}
    v \models \top                & \top                                        \\
    v \models \bot                & \bot                                        \\
    v \models x \sim k            & v(x) \sim k                                 \\
    v \models \neg \psi           & v \not\models \psi                          \\
    v \models \psi_1 \land \psi_2 & (v \models \psi_1) \land (v \models \psi_2)
  \end{tabular}
\end{center}

% \begin{displaymath}
%   \begin{array}{lcl}
%     s \models \bot                   & \leftrightarrow & \false                                               \\
%     v \models x \sim k               & \leftrightarrow & v(x) \sim k                                          \\
%     v \models \neg \psi              & \leftrightarrow & v \not \models \psi                                  \\
%     v \models \psi_1 \wedge \psi_{2} & \leftrightarrow & v \models \psi_{1} \textrm{ and } v \models \psi_{2} \\
%   \end{array}
% \end{displaymath}

%We say that a state \(s\) \emph{models the predicate \(\psi\)}, denoted by \(s \models \psi\), iff
%\(\psi\) evaluates to true when each \(x_i\) in the predicate is substituted with the
%value in the \(i^\text{th}\) dimension of \(s\).
%For some \(\psi \in \Pc(X)\), let \(\ValueSet{\psi}\) denote the
%set of valuations in \(S\) that satisfy \(\psi\), i.e., \(\ValueSet{\psi} =
%\Set*{s \given s \models \psi}\).

% \begin{definition}[Value-Predicate Distance~\cite{jaksic2018algebraic}]
%   Given a valuation $v$ and a predicate $\psi$ over $X$, we define the
%   \emph{value-predicate distance} as the Hausdorff Manhattan
%   distance between $v$ and the valuations satisfying $\psi$.
%
%   \begin{equation}
%     \vpd(v, \psi) = \min_{v' \models \psi} \sum_{x \in X} |v(x) - v'(x)|.
%   \end{equation}
% \end{definition}

In the following definition, we assume that there is an appropriately defined distance
metric $d$ between two valuations $v$ and
$v'$ (for example Manhattan distance as defined in
\autoref{eq:man}).
\begin{definition}[Value-Predicate Distance~\cite{jaksic2018algebraic}]
  Given a predicate \(\psi \in \Psi(X)\) and a valuation $v$, we define the
  \emph{value-predicate distance} as the distance between $v$ and the set of valuations
  that satisfy $\psi$ as follows:
  \begin{equation}
    \vpd(v, \psi) = \min_{v' \models \psi} d(v,v'),
  \end{equation}
  where \(d: D^n \times D^n \to \Re_{\geq 0}\) is a distance metric on the space of valuations.
\end{definition}

\begin{definition}[Symbolic Automaton~\cite{dantoni2017power}]
  A \emph{symbolic automaton} is a tuple \(\Ac = (X, Q, \init{q}, F, \Delta, \guards)\),
  where \(X\) is a finite set of variables, where each variable takes values in $D$;
  \(Q\) is a finite set of locations with initial location \(\init{q}\); \(F \subseteq
  Q\) is a set of \emph{accepting} locations; \(\Delta \subseteq Q \times Q\) is a
  nonempty set of transitions; and \(\guards: Q \times Q \to \Psi(X)\) is the guard
  labeling the transition.
\end{definition}
% \jyo{$\hat{q}$ is a bit weird for initial state.
%   Maybe $q_0$ or $q_\mathrm{init}$?
%   Best make it a macro.
% }
% \dn{I prefer $q_\mathrm{init}$, we might need $q_{0}$ when we just define a sequence of
%   locations}.

A \textit{run} of the symbolic automaton is defined as a sequence of states
and valuations for
variables in $X$ as follows:
\[
  q_0 \xrightarrow{v_1} q_1 \to \ldots \to  q_{n-1} \xrightarrow{v_n} q_n.
\]
Here, $q_0 = \init{q}$, and for all $i \in [0,n-1]$:
$(q_i,q_{i+1}) \in \Delta$, and
$v_{i+1} \models \guards(q_i,q_{i+1})$.
We say that the run is accepting if for some $n$,
$q_n \in F$.
We say that a symbolic automaton \(\Ac\) is \emph{terminally accepting} if for
every accepting state in $F$, all outgoing transitions are to some
state in $F$.
Such an automaton allows us to replace all accepting states by a single ``sink''
accepting state $q_F$, s.t.
$\forall (q_F,q) \in \Delta$, $q = q_F$.
In this paper, we restrict our attention to such terminally accepting symbolic automata.

\subsection{Reinforcement Learning}

%Given a set \(\Sigma\), we let \(\Dc(\Sigma)\) denote the set of probabilistic
%distributions over \(\Sigma\).
%For a distribution \(\mu \in \Dc(\Sigma)\) over \(\Sigma\), let
%\(\Supp(\mu)\) denote the support of \(\mu\) in \(\Sigma\).
%We also let \(2^\Sigma\) denote the power set of \(\Sigma\).

Reinforcement learning (RL) is a technique for an autonomous agent to learn the
\emph{policy} that maximizes some notion of a cumulative expected reward
provided to it by a stochastic environment.
Typically, the interaction between the RL agent and its environment is modeled as
Markov Decision Process (MDP).
% To formally define an MDP, we first introduce some notation.
% For a probability distribution $\theta \in \Dc(Y)$, we
% define the support of $\theta$, $\support(\theta) = \{y \mid \Pr(y) > 0\}$.
% A Markov Decision Process (MDP) is defined as follows:

\begin{definition}[Markov Decision Process (MDP)~\cite{sutton2018reinforcement}]
  An MDP is a tuple \(\Mc = (S, \init{s}, A, P, R)\), where $S$ is a finite set of
  states with initial state \(\init{s}\); \(A\) is a finite set of
  possible actions; \(P : S \times A \times S \to [0, 1]\) is a (partial) probabilistic
  transition function, where \(P(s, a, s') = \Pr\of{s' \given s, a}\) defines the
  probability of arriving in state \(s'\) after taking action \(a\) from state \(s\);
  and \(R: S \times S \rightarrow \Re\) is a reward function defined on
  \(\Mc\), where \(R(s,s')\) denotes the immediate reward received by transitioning
  from \(s\) to \(s'\).
\end{definition}
% \jyo{$\hat{q}$ looks even more weird now that we are using $s_0$ for MDP.}

An \emph{episode} \(\Sig = (s_0, \ldots, s_N)\) is a trace of length \(N\) in
the MDP \(\Mc\) such that \(s_0 = \init{s}\) and for all \(t
\in [0, N-1]\), \(P(s_t, a_t, s_{t+1}) > 0\) for some
\(a_t \in A\), and \(N\) is the maximum episode length.

Given a set \(Y\), we let \(\Dc(Y)\) denote the set of all probability
distributions over $Y$.
\begin{definition}[Policy of an MDP]
  A policy \(\pi: S \to \Dc(A)\) is a function that maps a state \(s \in S \) to
  a probability distribution over the set of actions $\Dc(A)$.
\end{definition}

Fixing a policy \(\pi\) in \(\Mc\) induces a probability space of episodic trajectories
characterized by the distribution \(\Mc^\pi\) such that the probability of generating a
trajectory \(\Sig\) in \(\Mc\) under the policy \(\pi\) (denoted \(\Sig \sim
\Mc^\pi\)):
\begin{displaymath}
  \Pr(\Sig \sim \Mc^\pi) = \Pr\of[\Big]{\left(s_0,\ldots,s_N\right) \given s_0 = \init{s}},
\end{displaymath}
where each action \(a_t\) is sampled from the distribution \(\pi(s_t)\),
and \(P(s_t,
a_t, s_{t+1}) > 0\).
\footnotemark
\footnotetext{We will remove the conditions \(a_t \sim \pi(s_t)\) and
  \(P(s_t, a_t, s_{t+1}) > 0\) where doing so is not ambiguous.
}

% Let \(\ReturnFn(\Sig)\) denote the \emph{total return} of a trajectory \(\Sig
% = (s_0,
% \ldots, s_N)\):
% \begin{equation}
%   \label{eq:total-returns}
%   \ReturnFn(\Sig) = \sum_{t=0}^{N-1} \Bof[\Big]{R(s_t, a_t, s_{t+1}) \given a_t \sim \pi(s_t), P(s_{i}, a_{i}, s_{i+1}) > 0}
% \end{equation}

%Let $s_{t}$ and $s_{t+1}$ denote previous and current state of the MDP with current time instance $t+1$. 
Let $R_t$ denote the immediate reward given to the agent in time
instance $t$ when the MDP transitions from state
$s_{t-1} = s$ to $s_{t} = s'$ : $R_t = R(s_{t-1}, s_{t}) = R(s,s')$.

\begin{definition}[Policy Value functions~\cite{sutton2018reinforcement}]
  Under a policy \(\pi : S \to \Dc(A)\), the \emph{state-value function} \(V^\pi: S \to
  \Re\) of some state \(s \in S\) in time instance \(1 \leq t \leq N\) is the expected
  total reward induced in \(\Mc^\pi\) starting from state \(s\):
  \begin{equation}
    \label{eq:value-function}
    V^\pi(s) = \Exp_{\pi} \Bof*{\sum_{i = t}^{N} R_{i}~|~s_{t} = s}
  \end{equation}

  Moreover, the \emph{action-value function} under a policy $\pi$, \(Q^\pi: S \times A
  \to \Re\) is the expected total reward for taking an action \(a \in A\) at some state
  \(s \in S\) at time-step \(t\):
  \begin{equation}
    \label{eq:q-function}
    Q^\pi(s, a) = \Exp_{\pi}\Bof*{\sum_{i = t}^{N} R_{i}~|~s_{t} = s,~a_{t}= a}
  \end{equation}
\end{definition}

The \emph{optimal} state-value function \(V^*(s)\) is defined as
\begin{displaymath}
  V^*(s) = \max_\pi V^\pi(s),~\forall s \in S,
\end{displaymath}
and the \emph{optimal} action-value function \(Q^*(s, a)\) is
\begin{displaymath}
  Q^*(s,a) = \max_\pi Q^\pi(s, a),~\forall s \in S, \forall a \in A.
\end{displaymath}
Note that the optimal policy value functions have recursive definitions:
\begin{align*}
  V^*(s) & = \max_{a \in A} Q^*(s,a)                                       \\
         & = \max_{a \in A} \Exp_{P(s, a, s') >0}\Bof*{R(s, s') + V^*(s')}
\end{align*}

The goal in reinforcement learning is to find a policy \(\pi^*\) on \(\Mc\) such that
\(V^{\pi^*}(\init{s}) =
V^*(\init{s})\), or
\begin{equation}
  \label{eq:optimal-policy}
  \pi^* = \argmax_{\pi} V^\pi(\init{s})
\end{equation}

% The goal in reinforcement learning is to find a policy \(\pi~:~S \to
% \Dc(A)\) on \(\Mc\) such that sampling actions from \(\pi\) maximizes
% the total reward expected on the probability space defined by the distribution
% $\Mc^\pi$.
% Formally, the RL problem involves solving the following optimization problem:
% \begin{displaymath}
%   \pi^{*}~=~\argmax_{\pi} V^{\pi}(\init{s}),
% \end{displaymath}
% where the \emph{value function} \(V^{\pi}(s)\) is the expected
% total reward
% obtained from taking actions sampled from the policy \(\pi\):
% \begin{displaymath}
%   V^\pi(s) = \Exp \left[ \sum_{t = 0}^{N-1} R(s_{t}, a_{t}, s_{t+1}) \middle| s_0 = s,
%     a_{t} \sim \pi(s_{t}), P(s_{t}, a_{t}, s_{t+1}) > 0 \right].
% \end{displaymath}

\begin{remark}
  Usually, the policy synthesis problem is posed for discounted, infinite
  runs~\cite{sutton2018reinforcement}.
  In this paper, we only consider the
  \emph{episodic reinforcement learning} setting~\cite{bertsekas1996neurodynamic}, where
  the goal is to maximize expected total returns from trajectories with a finite time
  bound \(N\) and initial state \(\init{s}\).
\end{remark}

\section{Problem Statement}
\label{sec:problem_s}

Given a Markov Decision Process \(\Mc = \Tuple{S, \init{s}, A, P}\), and a simply
accepting symbolic automaton specification \(\Ac = \Tuple{X, Q, \init{q}, F, \Delta, \guards}\) that
represents a task that needs to be performed in \(\Mc\), our goal is to synthesize a
policy that maximizes the probability of satisfying the task \(\Ac\).
To do this, we first define a product transition system which is an MDP that composes
\(\Mc\) with \(\Ac\):

\begin{definition}[Product MDP with accepting states]
  \label{def:product mdp}
  Given an MDP \(\Mc = \Tuple{S,\init{s}, A, P}\) with \(S \subset \Re^n\) and a
  symbolic automaton \(\Ac = \Tuple{X, Q, \init{q}, F, \Delta, \guards}\) with a
  valuation function \(v: S \times X \to D \) (where \(D \subset \Re\)), we can
  construct a product MDP (with additional annotation of accepting states) \(\Prod = \Mc
  \otimes \Ac\) as a tuple \(\Tuple{\prd{S},\prd{\init{s}},A,\prd{P}, \Acc}\), where:
  \begin{itemize}
    \item \(\prd{S} = S\times Q\),
    \item \(\prd{\init{s}} = (\init{s},\init{q})\),
          % \item \(\prd{A} = A\),
    \item \(\prd{P}: \prd{S} \times A \times \prd{S}
          \to [0,1]\) is defined as:
          \begin{displaymath}
            \begin{lgathered}
              \prd{P}((s,q),a,(s,q')) \\
              =
              \begin{cases}
                P(s,a,s') & ~\text{if}~ (q,q') \in \Delta, s \models \guards(q,q'), \\
                0         & ~\text{otherwise}.
              \end{cases}
            \end{lgathered}
          \end{displaymath}
    \item \(\Acc = \Set*{(s,q) \given q \in F}\).
  \end{itemize}
\end{definition}

An episode \(\Sig = ((s_0, q_0), \ldots, (s_N, q_N))\) in \(\Prod\) (with \((s_0, q_0)
= (\init{s}, \init{q})\)) is considered
\emph{accepting} if and only if \((s_N, q_N) \in Acc\).
We use \(\Sig \models \Ac\) to denote that the episode \(\Sig\) is accepted by the
specification automaton \(\Ac\).

Given a policy \(\pi: S \times Q \to \Dc(A)\), we let
\(\Pr(\pi \models \Ac)\) denote the probability that an episode sampled from
\(\Prod^\pi\) is accepted by \(\Ac\) (also called the \emph{probability of \(\pi\) being
  accepting}):
\begin{equation}
  \label{eq:prob-accepting}
  \begin{split}
    \Pr(\pi \models \Ac)
     & = \Pr_{\Sig \sim \Prod^\pi} \Bof[\big]{\Sig \models \Ac}               \\
     & = \Exp_{\Sig \sim \Prod^\pi} \Bof[\big]{\bm{1}\of*{\Sig \models \Ac}},
  \end{split}
\end{equation}
where \(\bm{1}(\cdot)\) is the indicator function such that
\begin{displaymath}
  \bm{1}(f) =
  \begin{dcases*}
    1, & if \(f\) evaluates to \(\top\) \\
    0, & otherwise.
  \end{dcases*}
\end{displaymath}

\begin{problem}
\label{problem:2}
Given an MDP \(\Mc = (S, \init{s}, A, P, R)\)  and a terminally accepting specification
automaton \(\Ac\), let \(\Prod = \Mc \otimes \Ac\).
Synthesize a policy \(\pi^*: S \times Q \to \Dc(A)\) that maximizes the probability of
acceptance.
\begin{displaymath}
  \pi^* = \argmax_{\pi} \Pr(\pi \models \Ac)
\end{displaymath}
\end{problem}

% \begin{lemma}
%   For any trajectory \(\hat{\xi} = (s_0, q_0), \ldots, (s_N, q_N)\) in the product
%   system \(\Prod\), the corresponding trajectory in \(\Mc\), \(\xi = s_0, \ldots, s_N\),
%   is \emph{accepting} if and only if \(\hat{\xi}\) is accepting, i.e., \((s_N, q_N) \in
%   \Acc\).
% \end{lemma}
% An alternative way to state this is that any policy \(\hat{\pi}\) in
% \(\Prod\) that is optimal with respect to
% \(\Pr(\hat{\pi} \models \Ac)\), we have that the same policy is also optimal in
% \(\Mc\)~\cite{courcoubetis1995complexity}.

% !TEX root = ../main.tex

\section{Rewarding Strategy For Symbolic Automata Goals}%
\label{sec:rewarding_strategy}

In the following sections, we will describe a rewarding strategy for the product
transition system that solves \problemname~\ref{problem:2} for the episodic
reinforcement learning setting.

For an MDP \(\Mc = (S, \init{s}, A, P, R)\), and a given specification
automaton \(\Ac = \Tuple{X, Q, \init{q}, F, \Delta, \guards}\), let \(\Prod = \Mc
\otimes \Ac = \Tuple{\prd{S}, \prd{\init{s}}, A, \prd{P}, \Acc}\) be the product MDP.
Then, we define \(R: \prd{S} \times \prd{S} \to \Re\) be
a reward function on \(\Prod\) such that:
\begin{equation}
  \label{eq:sparse-reward}
  R((s,q), (s',q')) =
  \begin{cases}
    \dmax, & ~\text{if}~ (s', q') \in \Acc    \\
           & ~\text{and}~ (s, q) \not\in \Acc \\
    % -\dmax, & ~\text{if}~ (s',q') \in \Rej  \\
    0,     & ~\text{otherwise}.
  \end{cases}
\end{equation}

Given the reward function \(R\), we claim that any policy \(\pi\) that maximizes the
expected total rewards using \(R\) will also maximize the probability of satisfying the
specification automaton \(\Ac\).
Formally:
\begin{theorem}
  \label{thm:sparse-reward-optimality}
  Let \(\pi_1\) and \(\pi_2\) be some policies on \(\Prod\) such that
  \(V^{\pi_1}((\init{s}, \init{q})) > V^{\pi_2}((\init{s}, \init{q}))\).
  Then, \(\Pr(\pi_1 \models \Ac) > \Pr(\pi_2 \models \Ac)\).
\end{theorem}
\begin{proof}
  For some trajectory \(\Sig = ((s_0, q_0), \ldots, (s_N, q_N))\) in \(\Prod^\pi\), let
  the total return for the trajectory be
  \begin{displaymath}
    \ReturnFn(\Sig) = \sum_{t=0}^{N - 1} R((s_t, q_t), (s_{t+1}, q_{t+1})).
  \end{displaymath}
  Since \(\Ac\) is terminally accepting, from~\autoref{eq:sparse-reward}, we can see
  that the set of all trajectories in \(\Prod\) can be partitioned into two sets:
  \begin{displaymath}
    \begin{array}{rcl}
      \Set*{\Sig \given G(\Sig) = \dmax} & ~\text{and}~ & \Set*{\Sig \given G(\Sig) = 0}
    \end{array}
  \end{displaymath}

  For notational convenience, we will use \(\init{V}^\pi\) (and \(\init{V}^*\)) denote
  the state-value function of policy \(\pi\) (and the optimal state-value function) for
  the initial state  \((\init{s}, \init{q})\) in \(\Prod\).

  For some policy \(\pi\), we know that
  \begin{align*}
    \init{V}^\pi
     & = \Exp_{\Sig \sim \Prod^\pi} G(\Sig)                                  \\
     & = \Exp_{\Sig \sim \Prod^\pi} \Bof*{\dmax \given \Sig \models \Ac}
    + \Exp_{\Sig \sim \Prod^\pi} \Bof*{0 \given \Sig \models \Ac}            \\
     & = \dmax \Exp_{\Sig \models \Prod^\pi} \Bof*{\bm{1}(\Sig \models \Ac)} \\
     & = \dmax \Pr(\pi \models \Ac)
  \end{align*}

  Thus, if \(\init{V}^{\pi_1} > \init{V}^{\pi_2}\), \(\Pr(\pi_1 \models \Ac) > \Pr(\pi_2
  \models \Ac)\).
\end{proof}

\begin{corollary}
  Let \(p^* = \max_\pi \Pr(\pi \models \Ac)\) be the maximum probability of acceptance
  of a policy \(\pi\)  in \(\Prod\), and let \(\pi^* = \argmax_\pi V^\pi ((\init{s},
  \init{q}))\) be an optimal policy with respect to the reward function \(R: \prd{S}
  \times \prd{S}\).
  Then, \(\Pr(\pi^* \models \Ac) = p^*\).
\end{corollary}

While the reward in~\autoref{eq:sparse-reward} provides theoretical guarantees for an
optimal accepting policy, this reward is \emph{sparse}, i.e., for large
episode lengths or task horizons, the agent may not see any rewards from accepting runs
in the early stages of its training.
This can cause considerable slowdown in the training process, and can potentially make
it unfeasible to use RL to synthesize such a controller.

To mitigate this, in the subsequent section will present a
\emph{reward~shaping}~\cite{grzes2017reward} technique that supplements spatial
information to \(R((s,q), (s',q'))\) and makes it \emph{dense}.

% \subimport{}{stefans_reward}

% !TEX root = ../main.tex

\subsection{Potential-based Reward Shaping}

To speed up the training process, we need to make the reward function \(R\) defined in
\autoref{eq:sparse-reward} more \emph{dense}, i.e., each transition in \(\Prod\)
needs to receive a reward such that:
\begin{enumerate}
  \item The reward is positive only if the agent moves closer to the goal;
  \item The shaped reward function does not alter the set of optimal policies.
\end{enumerate}
To this end, we define a \emph{potential-based reward shaping}
method~\cite{grzes2017reward} which uses a \emph{symbolic potential function} that
heuristically takes into account the shortest possible accepting trajectory from the
current state in \(\Prod\), {\em solely by looking at the symbolic constraints in} \(\Ac\).

We will show that this shaped reward function follows some basic requirements for
potential-based reward shaping such that any policy that optimizes this new reward
remains optimal under the reward defined in~\autoref{eq:sparse-reward}.

% To do this, we will first define a notion of progress in the task specification
% automaton \(\Ac = (X, Q, q_0, F, \Delta, g)\).
\begin{definition}[Task Progress Level, \(\eta\)]
  \label{def:eta}
  Given a terminally accepting automaton \(\Ac = (X, Q, q_0, F, \Delta, g)\), the
  \emph{task progress level} is a mapping \(\eta: Q \to \Ne \cup \{\infty\}\) such that
  \(\eta(q)\) is the length of the shortest simple path from the state \(q \in Q\) to
  the state \(q_F \in F\).
  If there is no such path from \(q\) to \(q_F\), then \(\eta(q) = \infty\).
\end{definition}

Let $\ValueSet{\psi(q,q')} = \Set*{s \given s\in S \wedge s \models \psi(q,q')}$ be the
set of all \(s \in S\) that satisfy the predicate \(\psi(q, q')\), and let
\(d_H\of*{\psi_1,\psi_2}\) be the Hausdorff distance between
\(\ValueSet{\psi_1}\) and \(\ValueSet{\psi_2}\) using some distance measure
\(d\) in \(S\).

\begin{definition}[Symbolic Subtask Progress, \(\PhiSym\)]
  The quantity \(\PhiSym: \Delta \to \Re_{\geq 0}\) is a \emph{heuristic}
  approximation of distance between a sub-goal set and the final goal set such that, for
  a transition \((q, q') \in \Delta\),

  \newlength\Normalbaselineskip
  \setlength\Normalbaselineskip{\baselineskip}
  \appto\MultlinedHook{\setlength\baselineskip{\Normalbaselineskip}}
  \begin{equation}
    \begin{array}{l}
      \PhiSym(q,q') = \\
      \begin{dcases}
        0 & \text{if}~ q' \in F                          \\
        \min\limits_{\substack{q'': (q', q'') \in \Delta \\
        q' \neq q''}}
        % \left(
        \begin{multlined}[t]
          d_H\of*{\ValueSet{\psi(q,q')}, \ValueSet{\psi(q', q'')}}\\
          \shoveleft{+ \PhiSym(q', q'')}
        \end{multlined}
        % \right)
          & \text{otherwise}.
      \end{dcases}
    \end{array}
  \end{equation}
\end{definition}

\begin{definition}[Symbolic Potential Function, \(\Phi\)] Thus, given an MDP \(\Mc =
  \Tuple{S, s_0, A, P}\) and a symbolic automaton task specification \(\Ac\), we define
  the function \(\Phi : S \times Q \to \Re_{\geq 0}\) in the product MDP \(\Prod = \Mc
  \otimes \Ac\) as:
  \begin{equation}
    \label{eq:true-potential}
    \Phi(s,q) =
    \begin{dcases}
      0 & \text{if}~(s,q) \in \Acc            \\
      \min_{\substack{q' : (q, q') \in \Delta \\
      \eta(q) \neq \eta(q') \lor              \\
      s \not\models \psi(q, q')}}
      % \left(
      \begin{multlined}[t]
        \vpd\of*{s, \psi(q, q')} \\ + \PhiSym(q, q')
      \end{multlined}
      % \right)
        & \text{otherwise}.
    \end{dcases}
  \end{equation}
\end{definition}

\begin{figure}[htbp]
  \begin{center}
    \includestandalone[width=\linewidth]{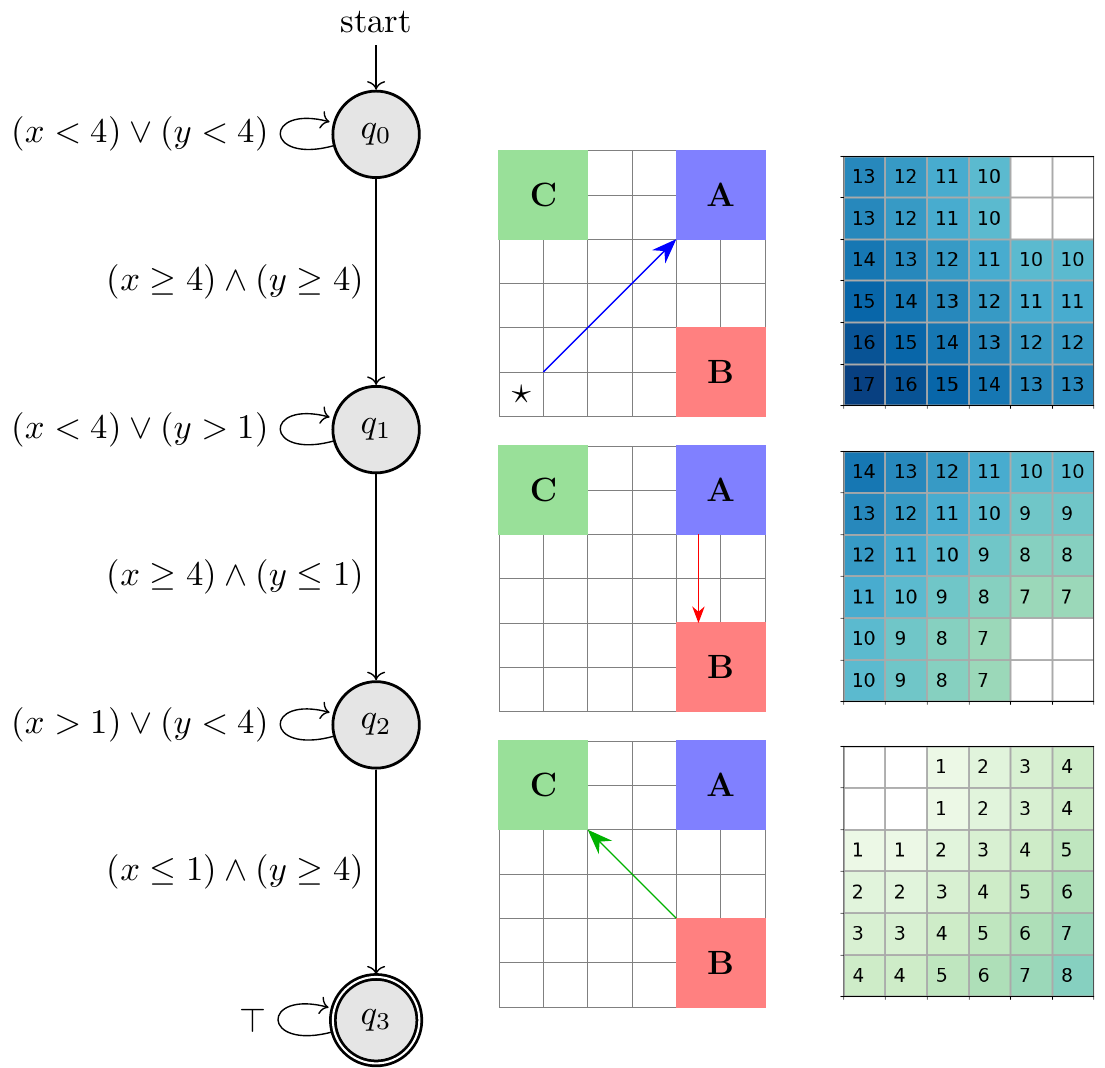}
  \end{center}
  \caption{%
    \textbf{Left:} The symbolic automaton for a sequential specification for the agent,
    where the goal is to visit regions \(A\), \(B\), and \(C\) in order.
    \textbf{Middle:} The approximate ``shortest path'' for the agent to satisfy the
    specification from \(\star\).
    \textbf{Right:} The evaluation of \(\Phi(s,q)\) from \autoref{eq:true-potential}
    such that the top image is for \(\Phi(s, q_0)\), the middle for \(\Phi(s, q_1)\),
    and the bottom for \(\Phi(s, q_2)\), for any \(s\).
  }%
  \label{fig:local_shaping}
\end{figure}

To gain some intuition behind how this potential function works, we refer
to~\autoref{fig:local_shaping}.
The quantity \(\eta(\cdot)\) allows us to check if we are making progress in
the specification automaton and disregard the transitions that don't contribute to the
symbolic progress measure \(\PhiSym(\cdot, \cdot)\).
\(\PhiSym(\cdot, \cdot)\) computes a symbolic approximation for the minimum length
path in the MDP that satisfies the specification.
For example, in the figure, given some state in \((s, q_0)\), making a transition to
some other state \((s', q_1)\), \(\PhiSym(q_0, q_1)\) is the minimum distance from
\((s, q_0)\) to the final goal state in \(\Acc\).
Thus, with a concrete value for \(s\), \(\Phi(s,q)\) computes the
approximate distance using the recursive formula.

% \begin{figure}[!htbp]
% \captionsetup[subfigure]{labelformat=empty}
%      \centering
%      \begin{minipage}{\textwidth}
%          \centering
%          \includegraphics[width=0.18 \textwidth]{figures/local_shaping_phi_q0.png}
%          %\caption{$\Phi(s, q_0)$}
%      \end{minipage}
%      %\hfill
%      \begin{minipage}{\textwidth}
%          \centering
%          \includegraphics[width=0.18\textwidth]{figures/local_shaping_phi_q1.png}
%          %\caption{$\Phi(s, q_1)$}
%      \end{minipage}
%      %\hfill
%      \begin{minipage}{\textwidth}
%          \centering
%          \includegraphics[width=0.18\textwidth]{figures/local_shaping_phi_q2.png}
%          %\caption{$\Phi(s, q_2)$}
%      \end{minipage}
%         \caption{The evaluation of $\Phi(s,q)$ from Eq. \eqref{eq:true-potential} on the example given in Figure \ref{fig:local_shaping}. Top-left $\Phi(s, q_0)$; Top-right $\Phi(s,q_1)$; Bottom-center $\Phi(s,q_2)$.}
%         \label{fig:phiplot}
% \end{figure}

% TODO: Define Length of a trajectory in the MDP as the total distance travelled by the agent in the MDP.
% 
% \begin{lemma}
%   \label{lemma:true-pot}
%   The function \(\Phi: S \times Q \to \Re_{\geq 0}\) is a potential function such that
%   for some \((s,q) \in \prd{S}\):
%   \begin{enumerate}
%     \item \(\Phi(s, q) = 0\) if and only if \((s,q) \in \Acc\);
%     \item For all \((s, q) \not\in \Acc\), \(\Phi(s,q) > 0\).
%   \end{enumerate}
%   
%   \begin{enumerate}
%       \item If \(\Phi(s,q) > \Phi(s',q')\) 
%   \end{enumerate}
% \end{lemma}
% \todo[author=anand,inline]{Proof.
%   ..}
% Assume \Phi(s,q) > 0, then:
% 
% 

From this definition, we can define the shaped reward function,
\(\hat{R}: \prd{S} \times
\prd{S} \to \Re\) as:
\begin{equation}
  \label{eq:shaped-reward}
  \begin{split}
    \hat{R}((s,q), (s',q')) & = R((s,q), (s',q'))             \\
                            & \quad+ \Phi(s,q) - \Phi(s',q'),
  \end{split}
\end{equation}
where \(R((s,q), (s',q'))\) is as defined in~\autoref{eq:sparse-reward}.

\begin{theorem}[Policy Invariance under Shaping~\cite{ng1999policy}]
  Let \(p^*\) be the maximum probability of acceptance.
  Let \(\pi\) be a policy that maximizes expected total rewards with respect to the
  sparse reward function \(R\), and \(\hat{\pi}\) be one that does so with the
  potential-based reward shaping \(\hat{R}\).
  Then, \(\Pr(\hat{\pi} \models \Ac) = \Pr(\pi \models \Ac) = p^*\).
\end{theorem}
\begin{proof}
  From~\cite{ng1999policy}, we know that any reward-optimal policy remains consistent
  with respect to the value function \(V^*((\init{s}, \init{q}))\) in \(\Prod\). Since
  any policy remains consistent under reward-shaping,
  from~\autoref{thm:sparse-reward-optimality}, we can see that the policy will remain
  optimal with respect to the probability of acceptance.
\end{proof}

% \todo[author=anand]{Can we just state that this arises from~\cite{ng1999policy}?}

In the subsequent section, we will empirically show the performance improvement gained
by using the proposed potential-based reward shaping method.

% !TEX root = ../main.tex

\section{Experiments}%
\label{sec:experiments}

\newcommand{\aUP}{\textsf{UP}}
\newcommand{\aDOWN}{\textsf{DOWN}}
\newcommand{\aLEFT}{\textsf{LEFT}}
\newcommand{\aRIGHT}{\textsf{RIGHT}}

In the following case studies, we consider an agent moving through a discrete, grid
environment, where the agent can use the actions \(A = \Set{\uparrow, \nearrow, \rightarrow, \searrow, \downarrow, \swarrow, \leftarrow, \nwarrow, 0}\) which
allow also for diagonal movements and no movement ($0$).
We model the probabilistic transition function \(P\) in the grid such that if the
controller decides to move along a direction, it will move to the next state (if there
is no wall) with probability \(1 - p_{\mathrm{slip}}\), or move along an
adjacent direction with probability \(0.5 p_\mathrm{slip}\).
Here, \(p_\mathrm{slip}\) is the probability of the agent ``slipping'', and is
set to \(0.1\).

\begin{figure}[H]
  \centering
  \includestandalone[width=.25\linewidth]{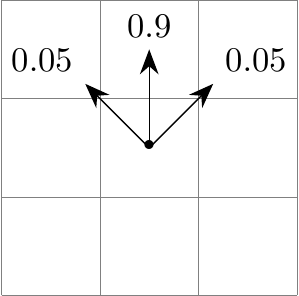}
  \caption{\aUP{} (\(\uparrow\)) transition for \(p_{slip} = 0.1\)}
\end{figure}

% \begin{figure}
%   %
%   \centering
%   \begin{minipage}{0.3\linewidth}%
%     \includestandalone[width=\linewidth]{map01}
%   \end{minipage}
%   %
%   \hfill%
%   \begin{minipage}{0.3\linewidth}%
%     \includestandalone[width=\linewidth]{map02}
%   \end{minipage}
%   %
%   \hfill%
%   \begin{minipage}{0.3\linewidth}%
%     \includestandalone[width=\linewidth]{map03}
%   \end{minipage}
%   %
%   \caption{
%     \label{fig:example}%
%     The types of tasks we will use to evaluate the efficacy of our method.
%     The first is a \emph{bounded reach} task, where the goal is to get from the initial
%     location \(\star\) to the goal set \(A\) within 15 time steps; the second is a
%     repeated satisfaction task where the agent has to visit the cells labeled \(A\) and
%     \(B\) every 4 timesteps for 16 timesteps; and the last is a sequential task where
%     the agent has to visit the regions labeled \(A\), \(B\), and \(C\) in order.
%   }%
% \end{figure}

We will compare the performance of our proposed symbolic potential-based reward
(\autoref{eq:shaped-reward}) against the sparse rewarding baseline
(\autoref{eq:sparse-reward}) and the potential-based reward presented
in~\cite{lavaei2020formal}.
We will also compare this with the performance of the purely quantitative approach
presented in~\cite{aksaray2016qlearning} to contrast some situations where a
quantitative method may outperform automata-based methods.

To do so, we will evaluate the training process for the tasks described below, and plot
the probability of acceptance for the learned policy at different points in the
training process.
At fixed intervals, we evaluate the policy 100 times, and aggregate the results across
5 training runs with different random seeds.
The probability of satisfaction is computed as the distribution of a binomial
distribution with 95\% confidence interval.

\begin{figure}
  \centering
  \begin{minipage}{0.35\linewidth}%
    \includestandalone[width=\linewidth]{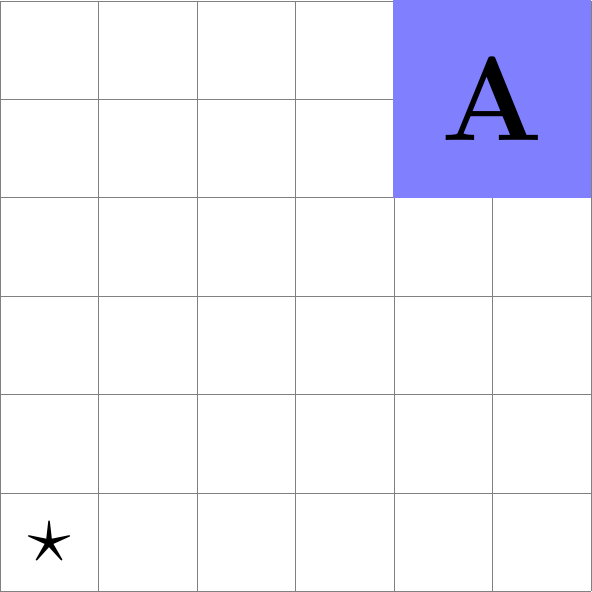}
  \end{minipage}
  \hfill
  \begin{minipage}{0.6\linewidth}%
    \includegraphics[width=\linewidth]{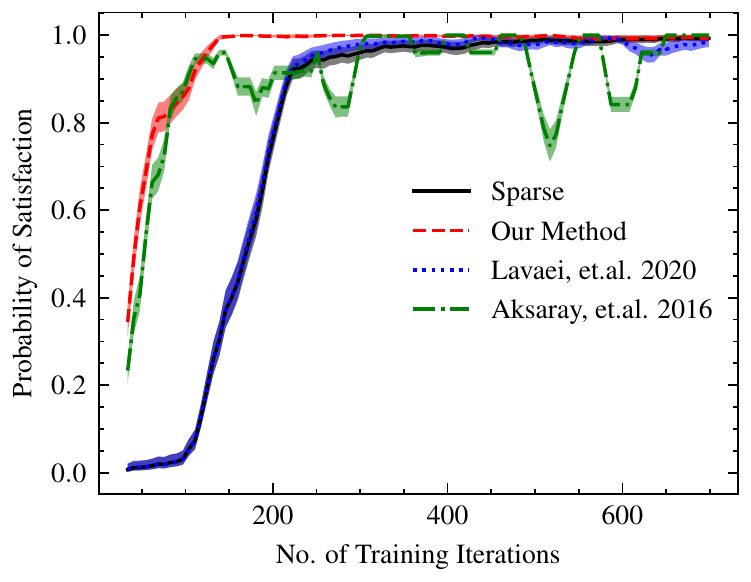}
  \end{minipage}
  % \hfill
  %
  \caption{%
    \textbf{Left}: The map for a simple ``reach'' task, where the goal is to get the
    agent from \(\star\) to the region labeled \(A\) within 14 time steps.
    \textbf{Right:} Plot of the probability of the learned policy generating an
    accepting trajectory vs.\ the training epoch.
  }%
  \label{fig:reach-results}%
\end{figure}

\paragraph{Reachability}
Here, the task of the RL agent is to start at some initial location \((s_0, q_0)\) and
reach some goal set \(A\) (represented in the automaton as a predicate).
Specifically, we are interested in synthesizing a controller in a \(6 \times
6\) grid environment, where the agent starts at state \((0,0)\) and needs to reach the
states satisfying \(x \geq 4 \land y \geq 4\) with a hard deadline of 15 time steps.
The efficacy of our approach can be seen in \figurename~\ref{fig:reach-results}.

From \autoref{fig:reach-results}, we can see that the our proposed method for reward
shaping is faster at finding a policy with acceptance probability equal to \(1.0\) than
other methods.
We see that the purely quantitative approach proposed in~\cite{aksaray2016qlearning} is
the next best solution, but suffers from poor stability in its results.
Moreover, in this scenario, the sparse rewarding strategy is exactly as performant as
the automaton potential-based reward shaping proposed in~\cite{lavaei2020formal}.
This is due to the fact that while a transition hasn't been taken in the automaton, the
potential function in~\cite{lavaei2020formal} provides no extra information.

\paragraph{Recurrence}

Based on an environment presented in~\cite{aksaray2016qlearning}, the goal of the agent
in this task is to repeatedly visit two regions in the map as often as possible within
a certain time limit.
Here, the goal is to visit two regions in a \(4 \times 4\) grid, labeled \(x = 2 \land
y = 2\) and \(x = 1 \land y = 3\).

\begin{figure}
  \centering
  \begin{minipage}{0.35\linewidth}%
    \includestandalone[width=\linewidth]{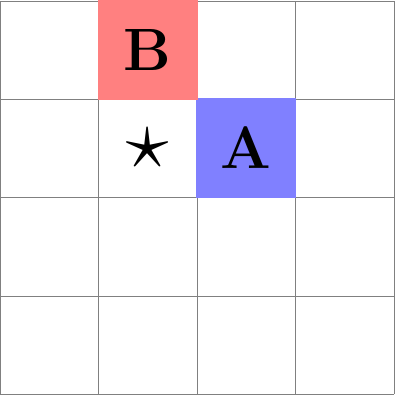}
  \end{minipage}
  \hfill%
  \begin{minipage}{0.6\linewidth}%
    \includegraphics[width=\linewidth]{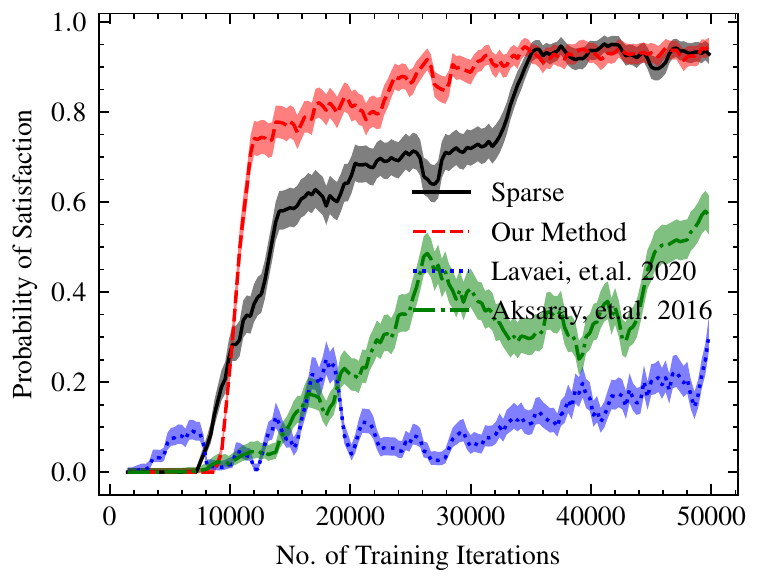}
  \end{minipage}
  \caption{%
    \textbf{Left:} The map for a ``recurrence'' or patrolling task, where the agent has
    to visit \(A\) and \(B\) within 5 time steps of each other over a span of 15 time
    steps.
    \textbf{Right:} Plot of the probability of the learned policy generating an
    accepting trajectory vs.\ the training epoch.
  }%
  \label{fig:recurrance-results}%
\end{figure}

In this environment we can see that the approach presented in~\cite{lavaei2020formal}
performs poorly.
This is due to the fact that in such specifications, using \(\eta(\cdot)\)
to compute the potential may mislead the agent into taking choices that are local
maximums in the rewards.
This is similar to the issue present in the ``Branching Paths'' task presented later.

On the other hand, increasing the task horizon (as defined
in~\cite{aksaray2016qlearning}) by a few time steps causes the \(\tau\)-MDP method to
perform poorly due to a state-space explosion.

\paragraph{Sequential}

This task requires an agent to visit regions in a strict sequence.
In this example, the agent is placed in a \(25 \times 25\) grid environment, with 3
labeled regions:
\begin{itemize}
  \item \(A = \Set*{(x, y) \given (x \geq 22) \land (y \geq 22)}\)
  \item \(B = \Set*{(x, y) \given (x \geq 22) \land (y \leq 3)}\)
  \item \(C = \Set*{(x, y) \given (x \leq 3) \land (y \geq 22)}\)
\end{itemize}
The goal of the agent is to learn a controller that visits the region A, then the
region B, and finally the region C in sequence.

\begin{figure}
  \centering
  \begin{minipage}{0.35\linewidth}%
    \includestandalone[width=\linewidth]{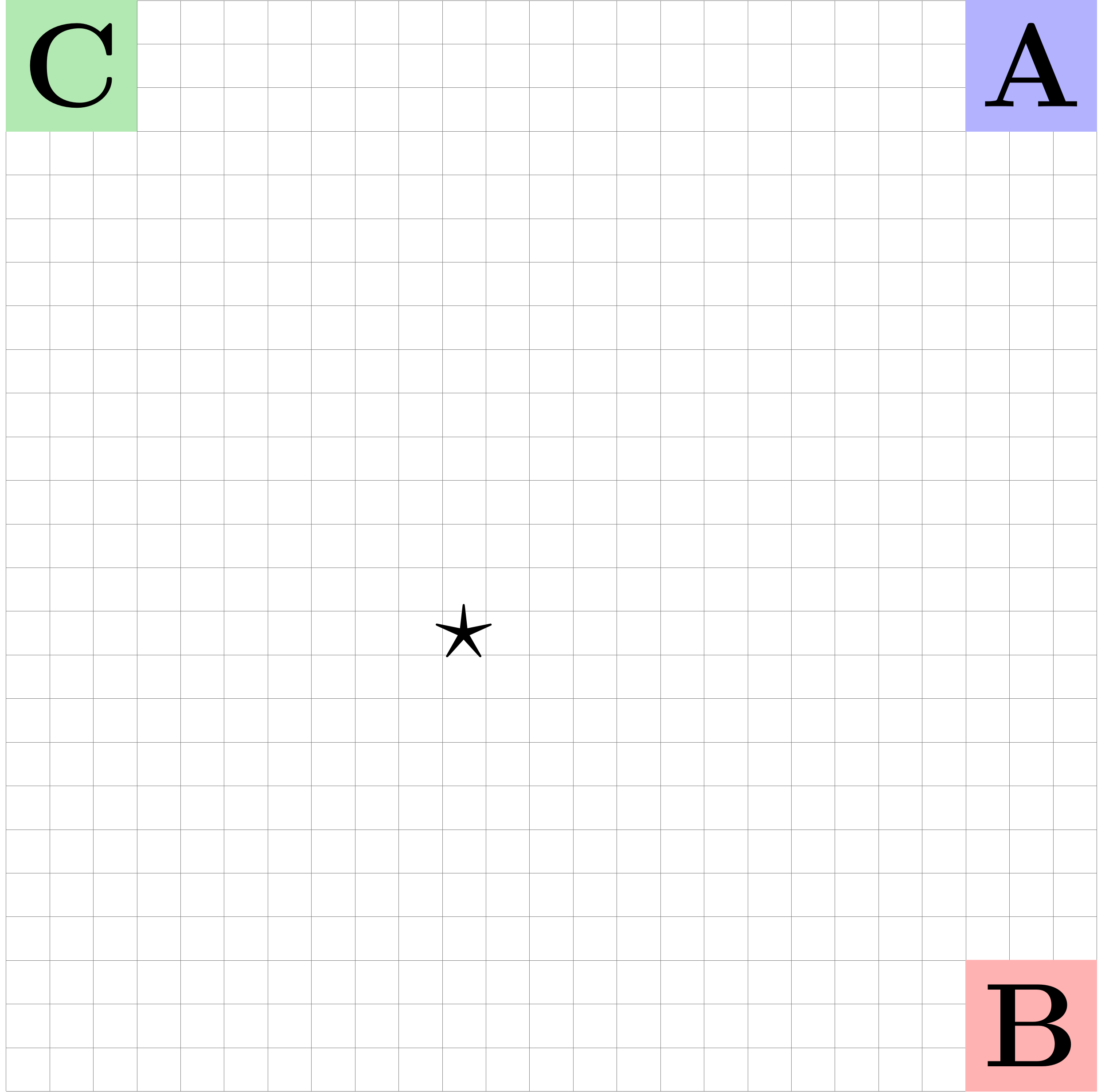}
  \end{minipage}
  \hfill%
  \begin{minipage}{0.6\linewidth}%
    \includegraphics[width=\linewidth]{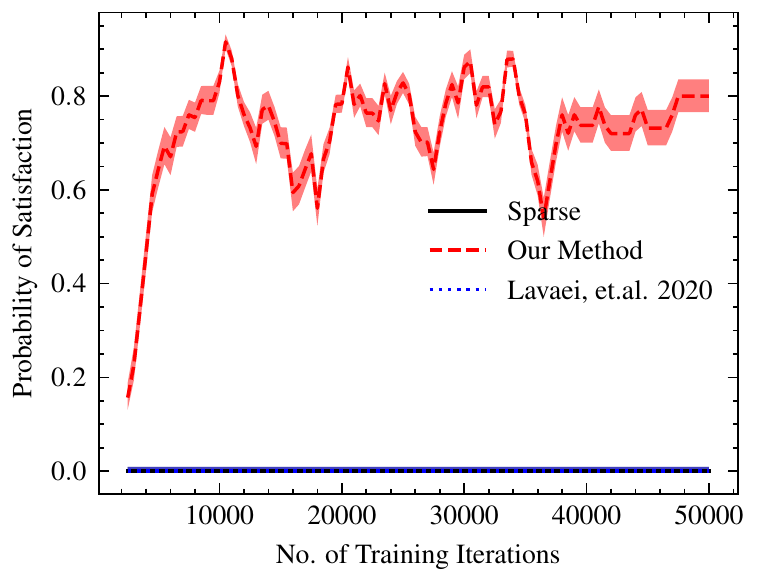}
  \end{minipage}
  \caption{%
    \textbf{Left:} The map for a simple sequential task, where the agent has
    to visit regions \(A\), \(B\), and \(C\) in that order.
    \textbf{Right:} Plot of the probability of the learned policy generating an
    accepting trajectory vs.\ the training epoch.
  }%
  \label{fig:sequential-results}%
\end{figure}

\begin{note}
  Since the environment and the task horizon are considerably large (even for
  artificially bounded task specifications), the state space for the
  \(\tau\)-MDP construct presented in~\cite{aksaray2016qlearning} explodes greatly.
  This made the experiments unfeasible to run with this method, and thus the results for
  this method are omitted from~\autoref{fig:sequential-results}.
\end{note}

In this task, we notice that the sparse reward baseline, along with the potential-based
reward shaping presented in~\cite{lavaei2020formal} do not learn any good information
for 50000 training epochs.
This is due to the incredibly sparse rewards provided by both
the methods.
Similar to the results in the ``Bounded Reach'' task presented earlier, the
method in~\cite{lavaei2020formal} does not provide any information to the agent until it
enters the region corresponding to the next task.

On the other hand, our proposed method learns to find satisfying traces relatively
quickly due to the information from the potential function
in~\autoref{eq:true-potential}.

\paragraph{Branching Paths}

\begin{figure}
  \centering
  \begin{minipage}{0.3\linewidth}%
    \includestandalone[width=\linewidth]{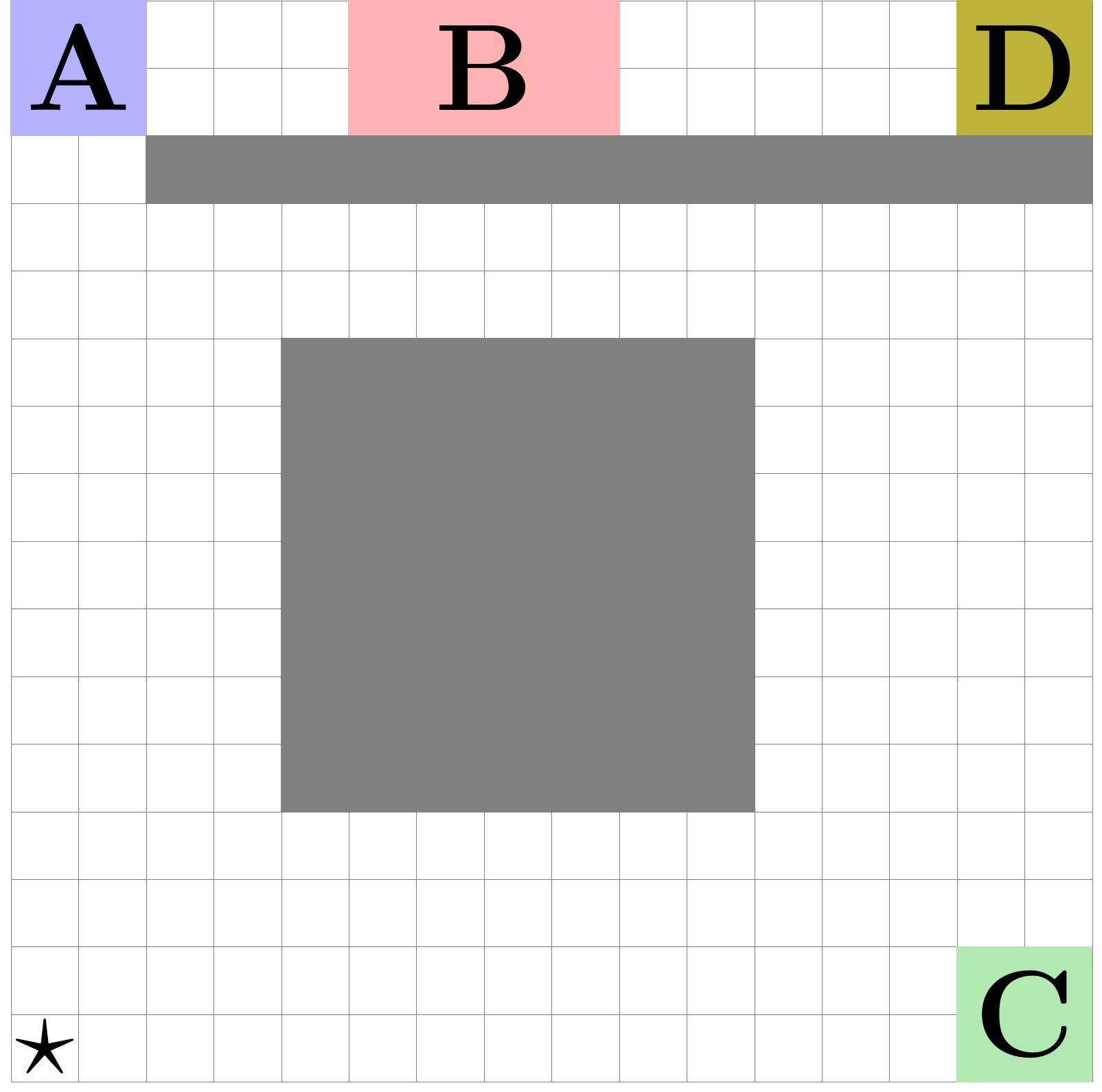}
  \end{minipage}
  \hfill%
  \begin{minipage}{0.6\linewidth}%
    \includegraphics[width=\linewidth]{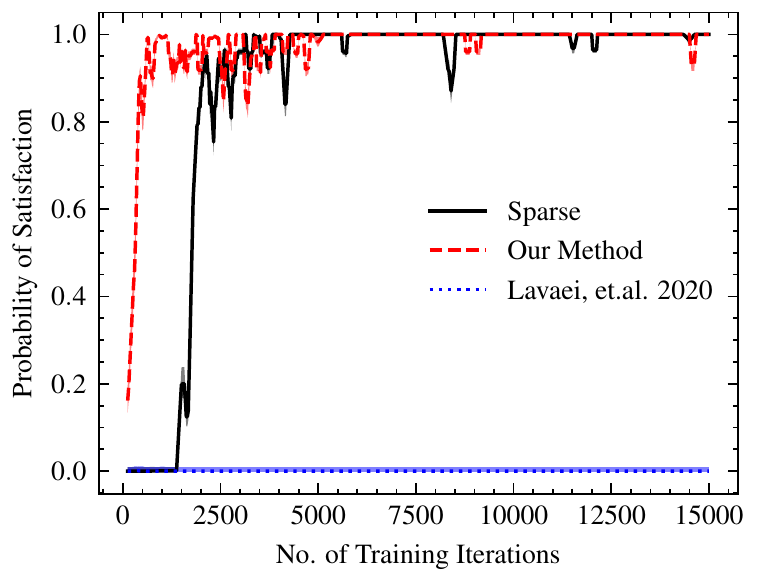}
  \end{minipage}
  \caption{%
    \textbf{Left:} The map for a task with two possible paths: the agent can either take
    the path \(A \to B \to D\) or \(C \to D\).
    Walls and obstacles have been intentionally placed to make one path easier than the
    other, but the agent does not have any knowledge of these environment features.
    \textbf{Right:} Plot of the probability of the learned policy generating an
    accepting trajectory vs.\ the training epoch.
  }%
  \label{fig:branched-reach-results}%
\end{figure}

In this specification, the agent operates on a \(16 \times 16\) grid with a few
obstacles, as seen in~\autoref{fig:branched-reach-results}.
The goal of the agent is to visit regions in either of the following orders: \(A \to B
\to D\) or \(C \to D\).
From the figure, we can see that one of the above orders is significantly longer than
the other, but since the agent does not have any prior knowledge of the environment
(except for the locations of these regions), it cannot rule out either branch.

\begin{note}
  For the same reasons as in the ``Sequential'' task, we do not evaluate the performance
  of the \(\tau\)-MDP approach presented in~\cite{aksaray2016qlearning}.
\end{note}

We can see from~\autoref{fig:branched-reach-results} that our proposed method
significantly outperforms the approach presented in~\cite{lavaei2020formal}.
(Note that due to size restrictions, we do not see that there is a small error band
around the graph for~\cite{lavaei2020formal} for when they do find some accepting
trajectories.)
Surprisingly, we see that the sparse rewarding case also performs well.
In the following section, we will analyze this result in detail and show that in this
case, the automaton structure can be misleading and lead the agent to local maximums in
the optimization problem.

% \begin{figure}
%   %
%   \centering
%   \begin{minipage}{0.45\linewidth}%
%     \includegraphics[width=\linewidth]{map01_results}
%   \end{minipage}
%   %
%   \hfill%
%   \begin{minipage}{0.45\linewidth}%
%     \includegraphics[width=\linewidth]{map02_results}
%   \end{minipage}
%   %
%   \hfill%
%   \begin{minipage}{0.45\linewidth}%
%     \includegraphics[width=\linewidth]{map03_results}
%   \end{minipage}
%   %
%   \caption{
%     \label{fig:results}%
%     The probability of acceptance of each task.
%   }%
% \end{figure}
%

\subsection{Case Study: Bounded Reach}%
\label{sub:bounded-reach}

\begin{figure}
  \begin{center}
    \includestandalone[width=0.75\linewidth]{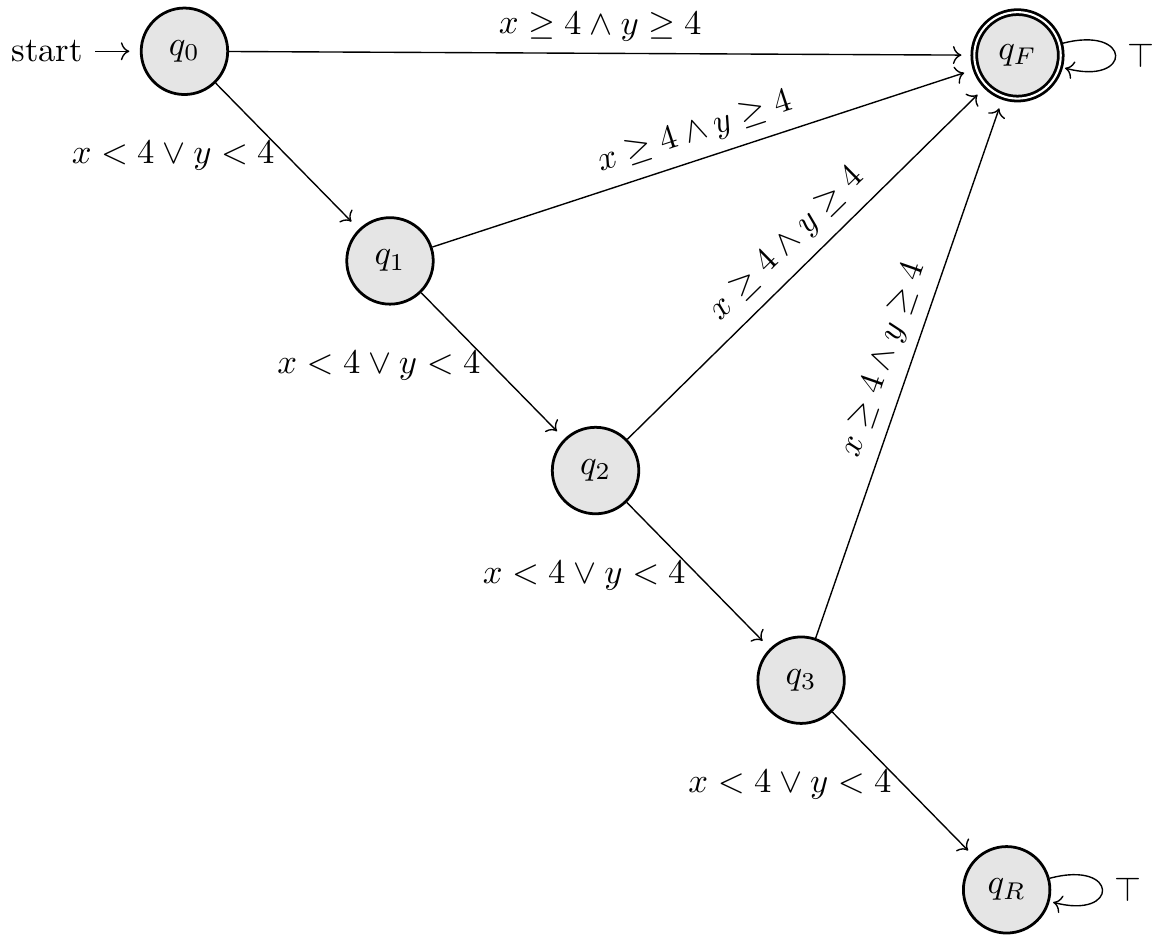}
  \end{center}
  \caption{This symbolic automaton represents the task ``reach the set represented by
    the constraints \(x \geq 4\) and \(y \geq 4\) within 4 time steps''.
    The state \(q0 = \init{q}\) is the initial state, \(q_F\) is the accepting final
    state, and the state \(q_R\) is the ``reject'' state, to which the system will
    transition if the agent fails to reach the goal state within the time bound.
    \(\)}%
  \label{fig:bounded-reach}
\end{figure}

First, we will look at the simple ``reach'' task, where the goal of the agent is to
reach a region \(A\) within some bounded time.
We will look at a simplified version of
the automaton in~\autoref{fig:bounded-reach}.
Here, we can see that for the states \(\Set{q_0, q_1, q_2, q_3}\), the value for
\(\eta(\cdot)\) is all equal, i.e., \(\eta(q) = 1\) for \(q
\in \Set{q_0, q_1, q_2,
  q_3}\).
Paraphrasing the potential-based reward function proposed
by~\cite{lavaei2020formal} below:
\begin{definition}[Automaton-based Reward Shaping~\cite{lavaei2020formal}]
  For an automaton \(\Ac = \Tuple{X, Q, \init{q}, F, \Delta, \guards}\), and the
  task progress level \(\eta: Q \to \Ne \cup \{\infty\}\),
  let \(\eta_{\max{}} = 1 + \max_{q \in Q} \Set{\eta(q) \given \eta(q) < \infty}\).
  Then,
  \begin{equation}
    \label{eq:lavaei-pot}
    \Phi_{\kappa}(q) =
    \begin{dcases}
      \kappa \frac{\eta(q) - \eta(\init{q})}{1 - \eta_{\max{}}} & \text{if}~ \eta(q) > 0
      \\
      1                                                         & \text{if}~ \eta(q) = 0,
    \end{dcases}
  \end{equation}
  where \(\kappa\) is a constant hyper-parameter.
  Then, the reward for a transition
  \(((s, q), (s', q'))\) in the product \(\Prod\) is:
  \begin{displaymath}
    \hat{R}_\kappa ((s,q), (s',q')) = \Phi_\kappa(q') - \Phi_\kappa(q)
  \end{displaymath}

\end{definition}

Notice from the above definition that while an agent doesn't satisfy the condition to
enter \(q_F\) in the bounded reachability problem in~\autoref{fig:bounded-reach}, the
reward obtained is \(0\), as \(\eta(q) - \eta(\init{q})\) for
all \(q \in \Set{q_0, q_1,
  q_2, q_3}\).
This causes the training process to be incredibly slow compared to our proposed
approach, where the \(\vpd(s, \cdot)\) term provides the agent with the
distance to the
goal set.

\subsection{Case Study: Branching Paths}%
\label{sub:branched-paths}

\begin{figure}[H]
  \begin{center}
    \includestandalone[width=\linewidth]{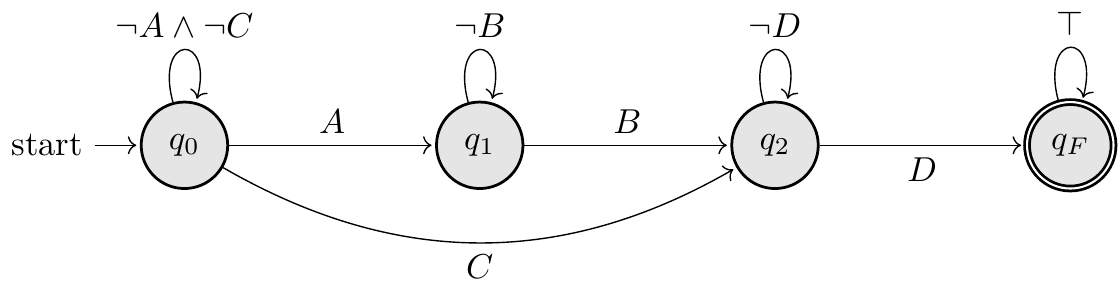}
  \end{center}
  \caption{%
    This symbolic automaton represents the task ``reach \(D\) by either going through
    \(A\) and then \(B\), or by going through \(C\)''.
    The state \(q_0 = \init{q}\) is the initial state, \(A = \left( (0 \leq x \leq 1)
    \land (14 \leq y \leq 15) \right)\), \(B = \left((5 \leq x \leq 8) \land (14 \leq
    y \leq 15) \right)\), and \(C = \left( (x >= 14) \land (y <= 1) \right)\).
  }%
  \label{fig:branch-reach}
\end{figure}

Here, we will look at the ``branching paths'' reachability task presented earlier,
represented by the symbolic automaton in \autoref{fig:branch-reach}.

Notice that for \(q \in \Set{q_0, q_1}\), \(\eta(q) = 2\), and
\(\eta(q_2) = 1\).
Thus, under the reward strategy presented by~\cite{lavaei2020formal} (see
\autoref{eq:lavaei-pot}), the agent will
get a positive reward only if it tries to take
either the \((q_1, q_2)\) transition into \(B\) or the \((q_0, q_2)\) transition into
\(C\).
This means that the agent will favor to go to \(C\) and then \(D\) due to the automaton
structure.
But, we can see from \autoref{fig:branched-reach-results} that picking the path to \(C\)
will cause the agent to take a much longer path than necessary.

We notice that this problem is mitigated when using our proposed potential function,
where both the paths in the automaton are equally favored (without the knowledge of the
obstacles in the environment). This, in turn, prevents the agent from getting stuck in
a local maximum. Similarly, in the sparse rewarding case, since all non-accepting paths
are equally weighted, this local maximum is still mitigated, compared to the approach
in~\cite{lavaei2020formal}.

\section{Related Work}%
\label{sec:related}

Reward engineering in RL based on formal specifications is a well-established research
topic~\cite{aksaray2016qlearning,hahn2019omegaregular,brafman2018ltlf,gaon2020reinforcement,venkataraman2020tractable,jothimurugan2021compositional,lavaei2020formal}.
In~\cite{aksaray2016qlearning} the authors define an effective approach for learning
robust controllers using Q-learning.
The history-based dependency of formula satisfaction is resolved by encoding n-step
history in every state.
The authors use bounded horizon robustness as a reward, which requires transforming MDP
by enhancing it with n-step MDP history.
A robustness-based approach to reward function was also taken in
~\cite{balakrishnan2019structured}.
In~\cite{lavaei2020formal}, the authors propose the use of a terminal deterministic
finite automaton (DFA) to encode task specifications over discrete labeled inputs, and
define a state-based potential function on the automaton.
Similarly, the authors of~\cite{jothimurugan2021compositional} propose the use of a
custom specification language~\cite{jothimurugan2019composable} to generate a similar
DFA, but rather than learning a single controller for the entire specification, they
propose to learn a controller for each ``subtask'' encoded on an edge in the form of
some guard.
These multiple controllers are then scheduled using the automaton structure as a guide.
Here we describe a more general method based on task progress, independent on the
specification language.
Our choice of automata as our specification language was inspired by a general
framework for calculating robustness w.r.t a specification expressed by a symbolic
weighted automaton~\cite{jaksic2018algebraic}.
The existence of direct translation from several specification formalisms to
automata~\cite{de2015synthesis,pnueli2008merits,bruggemann1993regular,bustan2005automata,jothimurugan2019composable} further supports our decision.

The problem of enabling Non-Markovian Rewards (NMRs) in RL can be understood as a
problem of supporting history-based behavior in an inherently memory-less MDP.
One of the approaches to supporting NMRs in RL is by introducing the Non-Markovian
Reward Decision Process (NMRDP), which allows reward functions to span over a history
of states~\cite{bacchus1996rewarding}.
Such NMRDPs can be transformed into a MDP which incorporates NMRs, previously
translated into automata~\cite{camacho2018non}.
In~\cite{gaon2020reinforcement}, we see that obtaining the NMR function by using
automata learning techniques comes at a cost of producing sufficient number of relevant
traces to feed the  learning algorithm.

A model-based technique for synthesizing MDP policies for LTL specifications was shown
in~\cite{kalagarla2020synthesis}, where the authors effectively reduce synthesis problem
to mixed integer linear programming.
A notion of task progress was used in~\cite{lacerda2015optimal} to generate policies for
Co-Safe LTL specifications.
A recent approach for LTL controller synthesis, which is also able to handle situations
when a task cannot be satisfied is presented in~\cite{guo2018probabilistic}.
A robustness-based method for model-free RL, called Temporal Logic Policy Search was
introduced in~\cite{li2018policy}.
Finally, in~\cite{alshiekh2018safe} the authors assign a ``shield" (based on a safety
automaton derived from LTL specifications) which helps guide the RL agent towards safe
(and more efficient) learning.

% !TEX root = ../main.tex

\section{Conclusion}%
\label{sec:conclusion}

In this paper, we present a novel approach to using symbolic automata as task specifications to encode complex tasks. We show that, compared to other automata-based solutions, the reward function obtained from this symbolic task specification can encode rich, quantitative information about the environment. We present theoretical guarantees for the correctness of the constructed reward, and empirically compare the approach against related works.

In future work, we hope to
\begin{enumerate*}
    \item extend these results in episodic reinforcement learning to infinite horizon tasks;
    \item provide guarantees for this approach in continuous space and continuous time settings; and
    \item study how to construct robust plans for multi-agent systems with global and local tasks.
\end{enumerate*}

% \nocite{*}
%% Use plainnat to work nicely with natbib. 
\bibliographystyle{plainnat}
\bibliography{bib}

\begin{thebibliography}{35}
\providecommand{\natexlab}[1]{#1}
\providecommand{\url}[1]{\texttt{#1}}
\expandafter\ifx\csname urlstyle\endcsname\relax
  \providecommand{\doi}[1]{doi: #1}\else
  \providecommand{\doi}{doi: \begingroup \urlstyle{rm}\Url}\fi

\bibitem[Aksaray et~al.(2016)Aksaray, Jones, Kong, Schwager, and
  Belta]{aksaray2016qlearning}
Derya Aksaray, Austin Jones, Zhaodan Kong, Mac Schwager, and Calin Belta.
\newblock {Q-Learning for Robust Satisfaction of Signal Temporal Logic
  Specifications}.
\newblock In \emph{2016 {{IEEE}} 55th {{Conference}} on {{Decision}} and
  {{Control}} ({{CDC}})}, pages 6565--6570, December 2016.
\newblock \doi{10.1109/cdc.2016.7799279}.

\bibitem[Alshiekh et~al.(2018)Alshiekh, Bloem, Ehlers, K~{\"o}nighofer, Niekum,
  and Topcu]{alshiekh2018safe}
Mohammed Alshiekh, Roderick Bloem, R{\"u}diger Ehlers, Bettina K~{\"o}nighofer,
  Scott Niekum, and Ufuk Topcu.
\newblock {Safe Reinforcement Learning via Shielding}.
\newblock \emph{Proceedings of the AAAI Conference on Artificial Intelligence},
  32\penalty0 (1), April 2018.
\newblock ISSN 2374-3468.
\newblock URL \url{https://ojs.aaai.org/index.php/AAAI/article/view/11797}.

\bibitem[Amodei et~al.(2016)Amodei, Olah, Steinhardt, Christiano, Schulman, and
  Man{\'e}]{amodei2016concrete}
Dario Amodei, Chris Olah, Jacob Steinhardt, Paul Christiano, John Schulman, and
  Dan Man{\'e}.
\newblock {Concrete Problems in AI Safety}.
\newblock \emph{arXiv:1606.06565 [cs]}, July 2016.
\newblock URL \url{http://arxiv.org/abs/1606.06565}.

\bibitem[Bacchus et~al.(1996)Bacchus, Boutilier, and
  Grove]{bacchus1996rewarding}
Fahiem Bacchus, Craig Boutilier, and Adam Grove.
\newblock {Rewarding behaviors}.
\newblock In \emph{Proceedings of the National Conference on Artificial
  Intelligence}, pages 1160--1167, 1996.

\bibitem[Balakrishnan and Deshmukh(2019)]{balakrishnan2019structured}
Anand Balakrishnan and Jyotirmoy~V Deshmukh.
\newblock {Structured reward shaping using signal temporal logic specifications
  }.
\newblock In \emph{2019 IEEE/RSJ International Conference on Intelligent Robots
  and Systems (IROS)}, pages 3481--3486. IEEE, November 2019.
\newblock \doi{10.1109/IROS40897.2019.8968254}.

\bibitem[Barto et~al.(1983)Barto, Sutton, and Anderson]{barto1983neuronlike}
Andrew~G. Barto, Richard~S. Sutton, and Charles~W. Anderson.
\newblock {Neuronlike Adaptive Elements That Can Solve Difficult Learning
  Control Problems}.
\newblock \emph{IEEE Transactions on Systems, Man, and Cybernetics},
  SMC-13\penalty0 (5):\penalty0 834--846, September 1983.
\newblock ISSN 0018-9472, 2168-2909.
\newblock \doi{10.1109/TSMC.1983.6313077}.

\bibitem[Bertsekas and Tsitsiklis(1996)]{bertsekas1996neurodynamic}
Dimitri~P. Bertsekas and John~N. Tsitsiklis.
\newblock \emph{{Neuro-Dynamic Programming}}.
\newblock Optimization and Neural Computation Series. {Athena Scientific},
  {Belmont, Mass}, 1996.
\newblock ISBN 978-1-886529-10-6.

\bibitem[Brafman et~al.(2018)Brafman, De~Giacomo, and Patrizi]{brafman2018ltlf}
Ronen Brafman, Giuseppe De~Giacomo, and Fabio Patrizi.
\newblock {LTLf/LDLf non-markovian rewards}.
\newblock In \emph{Proceedings of the AAAI Conference on Artificial
  Intelligence}, volume~32, 2018.

\bibitem[Br{\"u}ggemann-Klein(1993)]{bruggemann1993regular}
Anne Br{\"u}ggemann-Klein.
\newblock {Regular expressions into finite automata}.
\newblock \emph{Theoretical Computer Science}, 120\penalty0 (2):\penalty0
  197--213, 1993.

\bibitem[Bustan et~al.(2005)Bustan, Fisman, and Havlicek]{bustan2005automata}
Doron Bustan, Dana Fisman, and John Havlicek.
\newblock {Automata construction for PSL}.
\newblock Technical report, Citeseer, 2005.

\bibitem[Camacho et~al.(2018)Camacho, Chen, Sanner, and
  McIlraith]{camacho2018non}
Alberto Camacho, Oscar Chen, Scott Sanner, and Sheila~A McIlraith.
\newblock {Non-Markovian rewards expressed in LTL: Guiding search via reward
  shaping (extended version)}.
\newblock In \emph{GoalsRL, a workshop collocated with ICML/IJCAI/AAMAS}, 2018.

\bibitem[D'Antoni and Veanes(2017)]{dantoni2017power}
Loris D'Antoni and Margus Veanes.
\newblock {The Power of Symbolic Automata and Transducers}.
\newblock In Rupak Majumdar and Viktor Kun{\v c}ak, editors, \emph{Computer
  {{Aided Verification}}}, volume 10426, pages 47--67. Springer International
  Publishing, Cham, 2017.
\newblock ISBN 978-3-319-63386-2 978-3-319-63387-9.
\newblock \doi{10.1007/978-3-319-63387-9\_3}.
\newblock URL \url{https://link.springer.com/10.1007/978-3-319-63387-9\%5F3}.

\bibitem[De~Giacomo and Vardi(2015)]{de2015synthesis}
Giuseppe De~Giacomo and Moshe Vardi.
\newblock {Synthesis for LTL and LDL on finite traces}.
\newblock In \emph{Twenty-Fourth International Joint Conference on Artificial
  Intelligence}, 2015.

\bibitem[Gaon and Brafman(2020)]{gaon2020reinforcement}
Maor Gaon and Ronen Brafman.
\newblock {Reinforcement learning with non-markovian rewards}.
\newblock In \emph{Proceedings of the AAAI Conference on Artificial
  Intelligence}, volume~34, pages 3980--3987, 2020.

\bibitem[Grze{\'s}(2017)]{grzes2017reward}
Marek Grze{\'s}.
\newblock {Reward Shaping in Episodic Reinforcement Learning}.
\newblock In \emph{Proceedings of the 16th {{Conference}} on {{Autonomous
  Agents}} and {{MultiAgent Systems}}}, {{AAMAS}} '17, pages 565--573,
  {Richland, SC}, May 2017. {International Foundation for Autonomous Agents and
  Multiagent Systems}.

\bibitem[Guo and Zavlanos(2018)]{guo2018probabilistic}
Meng Guo and Michael~M Zavlanos.
\newblock {Probabilistic motion planning under temporal tasks and soft
  constraints}.
\newblock \emph{IEEE Transactions on Automatic Control}, 63\penalty0
  (12):\penalty0 4051--4066, 2018.

\bibitem[Hahn et~al.(2020)Hahn, Perez, Schewe, Somenzi, Trivedi, and
  Wojtczak]{hahn2020reward}
E.~M. Hahn, M.~Perez, S.~Schewe, F.~Somenzi, A.~Trivedi, and D.~Wojtczak.
\newblock {Reward Shaping for Reinforcement Learning with Omega- Regular
  Objectives}.
\newblock \emph{arXiv:2001.05977 [cs]}, January 2020.
\newblock URL \url{http://arxiv.org/abs/2001.05977}.

\bibitem[Hahn et~al.(2019)Hahn, Perez, Schewe, Somenzi, Trivedi, and
  Wojtczak]{hahn2019omegaregular}
Ernst~Moritz Hahn, Mateo Perez, Sven Schewe, Fabio Somenzi, Ashutosh Trivedi,
  and Dominik Wojtczak.
\newblock {Omega-Regular Objectives in Model-Free Reinforcement Learning}.
\newblock In Tom{\'a}{\v s} Vojnar and Lijun Zhang, editors, \emph{Tools and
  {{Algorithms}} for the {{Construction}} and {{Analysis} } of {{Systems}}},
  Lecture {{Notes}} in {{Computer Science}}, pages 395--412, Cham, 2019.
  Springer International Publishing.
\newblock ISBN 978-3-030-17462-0.
\newblock \doi{10.1007/978-3-030-17462-0\_27}.

\bibitem[Hasanbeig et~al.(2018)Hasanbeig, Abate, and
  Kroening]{hasanbeig2018logicallyconstrained}
Mohammadhosein Hasanbeig, Alessandro Abate, and Daniel Kroening.
\newblock {Logically-Constrained Reinforcement Learning}.
\newblock \emph{arXiv:1801.08099 [cs]}, January 2018.
\newblock URL \url{http://arxiv.org/abs/1801.08099}.

\bibitem[Jak{\v s}i{\'c} et~al.(2018)Jak{\v s}i{\'c}, Bartocci, Grosu, and
  Ni{\v c}kovi{\'c}]{jaksic2018algebraic}
Stefan Jak{\v s}i{\'c}, Ezio Bartocci, Radu Grosu, and Dejan Ni{\v c}kovi{\'c}.
\newblock {An Algebraic Framework for Runtime Verification}.
\newblock \emph{IEEE Transactions on Computer-Aided Design of Integrated
  Circuits and Systems}, 37\penalty0 (11):\penalty0 2233--2243, November 2018.
\newblock ISSN 1937-4151.
\newblock \doi{10.1109/TCAD.2018.2858460}.

\bibitem[Jothimurugan et~al.(2019)Jothimurugan, Alur, and
  Bastani]{jothimurugan2019composable}
Kishor Jothimurugan, Rajeev Alur, and Osbert Bastani.
\newblock {A Composable Specification Language for Reinforcement Learning
  Tasks}.
\newblock \emph{Advances in Neural Information Processing Systems}, 32, 2019.

\bibitem[Jothimurugan et~al.(2021)Jothimurugan, Bansal, Bastani, and
  Alur]{jothimurugan2021compositional}
Kishor Jothimurugan, Suguman Bansal, Osbert Bastani, and Rajeev Alur.
\newblock {Compositional Reinforcement Learning from Logical Specifications}.
\newblock \emph{Advances in Neural Information Processing Systems}, 34, 2021.

\bibitem[Kalagarla et~al.(2020)Kalagarla, Jain, and
  Nuzzo]{kalagarla2020synthesis}
Krishna~C Kalagarla, Rahul Jain, and Pierluigi Nuzzo.
\newblock {Synthesis of discounted-reward optimal policies for Markov decision
  processes under linear temporal logic specifications}.
\newblock \emph{arXiv preprint arXiv:2011.00632}, 2020.

\bibitem[Lacerda et~al.(2015)Lacerda, Parker, and Hawes]{lacerda2015optimal}
Bruno Lacerda, David Parker, and Nick Hawes.
\newblock {Optimal policy generation for partially satisfiable co-safe LTL
  specifications}.
\newblock In \emph{Twenty-Fourth International Joint Conference on Artificial
  Intelligence}, 2015.

\bibitem[Lavaei et~al.(2020)Lavaei, Somenzi, Soudjani, Trivedi, and
  Zamani]{lavaei2020formal}
Abolfazl Lavaei, Fabio Somenzi, Sadegh Soudjani, Ashutosh Trivedi, and Majid
  Zamani.
\newblock {Formal Controller Synthesis for Continuous-Space MDPs via Model-Free
  Reinforcement Learning}.
\newblock \emph{arXiv:2003.00712 [cs, eess]}, pages 98--107, March 2020.
\newblock \doi{10.1109/ICCPS48487.2020.00017}.
\newblock URL \url{http://arxiv.org/abs/2003.00712}.

\bibitem[Li et~al.(2018)Li, Ma, and Belta]{li2018policy}
Xiao Li, Yao Ma, and Calin Belta.
\newblock {A policy search method for temporal logic specified reinforcement
  learning tasks}.
\newblock In \emph{2018 Annual American Control Conference (ACC)}, pages
  240--245. IEEE, 2018.

\bibitem[Mnih et~al.(2015)Mnih, Kavukcuoglu, Silver, Rusu, Veness, Bellemare,
  Graves, Riedmiller, Fidjeland, Ostrovski, Petersen, Beattie, Sadik,
  Antonoglou, King, Kumaran, Wierstra, Legg, and Hassabis]{mnih2015humanlevel}
Volodymyr Mnih, Koray Kavukcuoglu, David Silver, Andrei~A. Rusu, Joel Veness,
  Marc~G. Bellemare, Alex Graves, Martin Riedmiller, Andreas~K. Fidjeland,
  Georg Ostrovski, Stig Petersen, Charles Beattie, Amir Sadik, Ioannis
  Antonoglou, Helen King, Dharshan Kumaran, Daan Wierstra, Shane Legg, and
  Demis Hassabis.
\newblock {Human-Level Control through Deep Reinforcement Learning}.
\newblock \emph{Nature}, 518\penalty0 (7540):\penalty0 529--533, February 2015.
\newblock ISSN 1476-4687.
\newblock \doi{10.1038/nature14236}.

\bibitem[Mnih et~al.(2016)Mnih, Badia, Mirza, Graves, Lillicrap, Harley,
  Silver, and Kavukcuoglu]{mnih2016asynchronous}
Volodymyr Mnih, Adria~Puigdomenech Badia, Mehdi Mirza, Alex Graves, Timothy
  Lillicrap, Tim Harley, David Silver, and Koray Kavukcuoglu.
\newblock {Asynchronous Methods for Deep Reinforcement Learning}.
\newblock In \emph{International {{Conference}} on {{Machine Learning}}}, pages
  1928--1937, June 2016.
\newblock URL \url{http://proceedings.mlr.press/v48/mniha16.html}.

\bibitem[Ng et~al.(1999)Ng, Harada, and Russell]{ng1999policy}
Andrew~Y Ng, Daishi Harada, and Stuart Russell.
\newblock {Policy invariance under reward transformations: Theory and
  application to reward shaping}.
\newblock In \emph{ICML}, volume~99 of \emph{{{ICML}} '99}, pages 278--287,
  {San Francisco, CA, USA}, June 1999. {Morgan Kaufmann Publishers Inc.}
\newblock ISBN 978-1-55860-612-8.

\bibitem[Pnueli and Zaks(2008)]{pnueli2008merits}
Amir Pnueli and Aleksandr Zaks.
\newblock {On the merits of temporal testers}.
\newblock In \emph{25 Years of Model Checking}, pages 172--195. Springer, 2008.

\bibitem[Sadigh et~al.(2014)Sadigh, Kim, Coogan, Sastry, and
  Seshia]{sadigh2014learning}
Dorsa Sadigh, Eric~S. Kim, Samuel Coogan, S.~Shankar Sastry, and Sanjit~A.
  Seshia.
\newblock {A Learning Based Approach to Control Synthesis of Markov Decision
  Processes for Linear Temporal Logic Specifications}.
\newblock In \emph{53rd {{IEEE Conference}} on {{Decision}} and {{Control}}},
  pages 1091--1096, December 2014.
\newblock \doi{10.1109/cdc.2014.7039527}.

\bibitem[Schulman et~al.(2017)Schulman, Wolski, Dhariwal, Radford, and
  Klimov]{schulman2017proximal}
John Schulman, Filip Wolski, Prafulla Dhariwal, Alec Radford, and Oleg Klimov.
\newblock {Proximal Policy Optimization Algorithms}.
\newblock \emph{arXiv:1707.06347 [cs]}, August 2017.
\newblock URL \url{http://arxiv.org/abs/1707.06347}.

\bibitem[Silver et~al.(2014)Silver, Lever, Heess, Degris, Wierstra, and
  Riedmiller]{silver2014deterministic}
David Silver, Guy Lever, Nicolas Heess, Thomas Degris, Daan Wierstra, and
  Martin Riedmiller.
\newblock {Deterministic Policy Gradient Algorithms}.
\newblock In \emph{{{ICML}}}, June 2014.
\newblock URL \url{https://hal.inria.fr/hal-00938992}.

\bibitem[Sutton and Barto(2018)]{sutton2018reinforcement}
Richard~S Sutton and Andrew~G Barto.
\newblock \emph{{Reinforcement learning: An introduction}}.
\newblock Adaptive Computation and Machine Learning Series. MIT press,
  {Cambridge, Massachusetts}, second edition edition, 2018.
\newblock ISBN 978-0-262-03924-6.

\bibitem[Venkataraman et~al.(2020)Venkataraman, Aksaray, and
  Seiler]{venkataraman2020tractable}
Harish Venkataraman, Derya Aksaray, and Peter Seiler.
\newblock {Tractable reinforcement learning of signal temporal logic objectives
  }.
\newblock In \emph{Learning for Dynamics and Control}, pages 308--317. PMLR,
  {PMLR}, July 2020.
\newblock URL \url{https://proceedings.mlr.press/v120/venkataraman20a.html}.

\end{thebibliography}

\end{document}